%% file: coling2020.tex
\documentclass[11pt]{article}
\usepackage{coling2020}
\usepackage{times}
\usepackage{url}
\usepackage{latexsym}
\usepackage{amsmath}
\usepackage{stmaryrd}
\usepackage{graphicx}
\usepackage{numprint}
\usepackage{multicol}%
\usepackage{multirow}%
\usepackage{tikz}
\usepackage{tkz-graph}
\usepackage{tkz-euclide}
\usepackage{subfig}
\usetikzlibrary{shapes,decorations}
\usepackage{relsize}
\usepackage{xcolor}
\usepackage{pgfplots}
\pgfplotsset{compat=newest}
\usepackage{pgfplotstable}
\usepgfplotslibrary{colorbrewer}
\definecolor{theblue}{rgb}{.2,.2,.7}
\definecolor{thegreen}{rgb}{.33, .66, .22}

\newcommand{\exInText}[1]{``\textit{#1}''}

\newcommand{\editsTM}[1]{#1}
\newcommand{\editsTB}[1]{#1}
\newcommand{\editsDP}[1]{#1}

\usepackage{csquotes} 

\colingfinalcopy 

\title{What Meaning-Form Correlation Has to Compose With: \\ A Study of \textsc{Mfc} on Artificial and Natural Language}

\author{Timothee Mickus%
  \thanks{\ \ Work done while at \textsc{Aist} (Japan).}
  \\
  \textsc{Atilf} \\
  {\tt tmickus@atilf.fr} \\\And
  Timothée Bernard%
  \footnotemark[1]
  \\
  Université de Paris \\
  {\tt timothee.bernard}\\{\tt @ens-lyon.org}  \\\And
  Denis Paperno \\
  Utrecht University \\
  {\tt d.paperno@uu.nl}\\}

\date{}

\begin{document}
\maketitle
\begin{abstract}
    Compositionality is a widely discussed property of natural languages, although its exact definition has been elusive.
    We focus on the proposal that compositionality can be assessed by measuring meaning-form correlation.
    We analyze meaning-form correlation on three sets of languages:
    (i) \editsTM{artificial} toy languages tailored to be compositional,
    (ii) a set of English dictionary definitions, and
    (iii) a set of English sentences drawn from literature.
    We find that \editsDP{linguistic phenomena such as synonymy and ungrounded stop-words weigh on \textsc{mfc} measurements}, 
    and that straightforward methods to mitigate their effects have widely varying results depending on the dataset they are applied to.
    \editsTM{Data and code are made publicly available.}
\end{abstract}

\section{Introduction}
\label{sec:intro}
\blfootnote{
    %
    %
    \hspace{-0.65cm}  
    This work is licensed under a Creative Commons 
    Attribution 4.0 International License.
    License details:
    \url{http://creativecommons.org/licenses/by/4.0/}.
}

Compositionality is one of the core aspects of language: when speaking, we weave complex propositions out of unanalyzable atoms of meaning.
This intricate interplay between symbols and sentences has been an important point of research
in areas ranging from  philosophy of language \cite{frege1884GdA} to distributional semantics \cite{Baroni2014FregeIS}, 
and has been held as one of the key characteristic differentiating human language from animal communication \cite{Hockett1960OrigSpeech}.
Compositionality in natural language is usually captured by the Principle of Compositionality.
This principle, often attributed to Frege although this paternity is debatable \cite{Hinzen12CompoCtxt}, has been worded in different ways; let us quote a classic textbook \cite[p.26]{gamut_logic_1991}: 
\enquote{the meaning of a composite expression must be wholly determined by the meanings of its composite parts and of the syntactic rule by means of which it is formed}.

One suggestion, that can be traced back to \newcite{kirby99syntaxOutOfLearning}, and that was fully operationalized in \newcite{kirby08cumul}, is that compositionality can be measured as a correlation between meaning and surface form \editsTB{(i.e., the sequence of tokens)}.
If we consider sentences such as \exInText{I saw a black cat} and \exInText{I saw a black crow}, we see that minute changes in the form of a sentence entail that its meaning does not change too much.
Compare these examples to sentences such as \exInText{I saw a white crow}, or \exInText{He saw a white crow}: the gradual alteration of the form of a sentence also gradually alters its meaning.
Provided that we can measure similarity between any two sentences forms, and between any two sentences meanings, this observation can be rephrased as follows: the distance between two sentences meanings should correlate with the distance between their forms.
We refer to this as \emph{meaning-form correlation}, or \textsc{mfc} for short.

One area where \textsc{mfc} has been employed as a way to both detect and quantify compositionality is the field of emergent communication \cite[a.o.]{kirby08cumul,Kirby2015CompressionAC,spike2017minimal-thesis,Ren2020Compositional} which studies agents (artificial or human) who have to produce messages in order to express well defined meanings. 
To estimate the compositionality of a set of message-meaning pairs, one can compute a Pearson or Spearman correlation score between textual distances and meaning distances.
Additionally, a permutation test (over random permutation of the message-meaning assignment) can produce a p-value indicating whether the correlation is statistically significant as well as a z-score that quantifies in number of standard deviations the difference between the correlation found in the language and the correlation of random assignments.
This is done among others by \newcite{Kirby2015CompressionAC}, \newcite{gutierrez-etal-2016-finding}, or \newcite{spike2017minimal-thesis} using a Mantel test \cite{mantel1967detection}, which is based on Pearson's notion of correlation. 

To the extent of our knowledge, no study has investigated the validity of this methodology. 
Is \textsc{mfc} a coherent method of measuring compositionality, especially when it comes to \emph{natural language}? Or do other factors weigh on the measurements and obfuscate the results?
If so, can we identify these factors, and can we control for them?
To address this gap, we therefore study how \textsc{mfc} behaves, both in a controlled setup based on \editsTM{artificial} data and in more realistic scenarios involving natural language.\footnote{%
    \editsTM{Data and code are publicly available at \url{https://github.com/TimotheeMickus/mf-correl}.}
}

\section{\textsc{Mfc} and \editsTM{Artificial L}anguages}
\label{sec:synthlang}

\textsc{Mfc}-based assessments of compositionality implicitly assume that any change in form should correspond to some change in meaning. 
\editsTM{
This can however be challenged: for instance, synonyms and paraphrases will introduce changes in form that should not entail change in meaning.} Thus we expect \textsc{mfc} to be sensitive to such phenomena, and this in turn suggests that factors such as synonymy could overpower the effect of compositionality that we wish to detect using \textsc{mfc}.
\editsTM{
To approach this question, we generate} \editsTM{artificial} languages containing varying degrees of compositionality as well as potential confounding factors\editsTM{.}

\subsection{Methodology} 

Our experimental protocol consists in generating \editsTM{artificial} languages with varying properties, and see how these properties impact \textsc{mfc} measurements.
All of our \editsTM{artificial} languages are sets of message-meaning pairs.
We represent meanings as binary vectors of five components, whereas messages are sequences of symbols. 
We refer to each of the five semantic dimensions as a \emph{concept}.
In most cases, the value of a concept will be denoted in a message by a specific symbol, which we call its \emph{expression}. 

As we are interested in whether \textsc{mfc} accurately captures \emph{compositionality}, we design languages where some of the concepts are systematically expressed conjointly by unanalyzable holistic expressions.
We generate languages where the values of the first $h$ concepts are systematically expressed through a single expression, and the other $5-h$ are left untouched, with $h$ varying from 1 to 5. When $h=1$, the language is entirely compositional; when $h=5$, the language is entirely holistic.


We also consider \emph{synonymy} as a possible confounding factor: as previously noted, synonyms entail a variation in form that is not coupled with a variation in meaning.
To model this phenomenon, we generate languages in which any single value of a concept can equally be expressed by $s$ different expressions, with $s$ ranging from 1 to 3. 
As a consequence, when $s=1$, the language exhibits no synonymy.


\editsTM{Moreover, we expect that \textsc{mfc} measurements might be impacted by the presence of \emph{semantically ungrounded elements}---i.e., elements not associated with any concept or combination of concepts. }
We therefore generate languages where $u$ specific ungrounded symbols appear once in every message at randomly chosen positions, with $u$ varying between 0 and 3.
In languages where $u=0$, the language contains only semantically grounded expressions.

Finally, we consider the case of \emph{paraphrases}, sentences of different forms but equivalent meanings.
For a language to contain paraphrases, it must be able to express a single meaning with different messages. 
This is the case in our \editsTM{artificial} languages that exhibit synonymy or contain semantically ungrounded elements: the variation they introduce allow distinct messages to have the same semantic.
We thus generate languages for which $p$ messages are produced for each meaning before dropping possible meaning-message pair duplicates. 
$p$ ranges from 1 to 3.
If $p=1$, the language contains no paraphrase.

We test all possible combinations of these four parameters. 
We also include \emph{random baselines} where we assign meanings to \editsTM{random sequences of symbols,} 
either of an arbitrarily fixed length of 5 symbols, or of a length chosen uniformly between 1 and 10.
%
We generate 50 \editsTM{artificial} languages for every combination of parameters 
\editsTM{to} help us distinguish the stable effects of our parameters from spurious accidents due to our random generation process. We refer to each of the 50 generation processes as a separate \emph{run}. 
\editsTM{We 
compute Mantel tests using 
the Hamming distance between meaning vectors---i.e., the number of differing components---and 
the Levenshtein distance over messages, normalized by the maximum length of the two messages.}
For each language, we study the corresponding p-value and the correlation score\editsTM{.}
For every combination of parameters, we study its average p-value and correlation score across \editsTM{all runs.} 



\editsTM{One limitation of this methodology is that our modeling may not comply fully with natural language---in particular, the existence of exact synonyms is debatable; likewise, natural function words do possess some semantic content, whereas our ungrounded symbols do not. Neither do we claim to conduct an exhaustive study of all relevant phenomena: we leave other likely factors such as multi-word expressions to future work.}

\subsection{Results} 
\begin{figure}[t]
    \centering
    \subfloat[\label{fig:synth-langs:hol}Compositionality \newline{\tiny NB: No $h=5$ is found significant.}]{\includegraphics[scale=0.4]{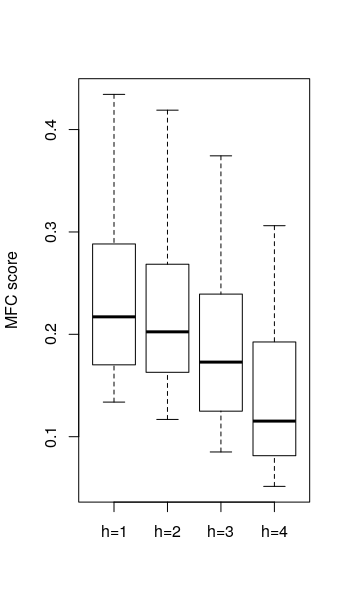}} \quad
    \subfloat[\label{fig:synth-langs:syn}Synonymy]{\includegraphics[scale=0.4]{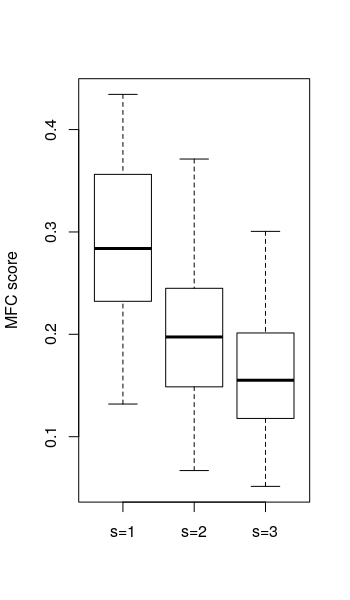}} \quad
    \subfloat[\label{fig:synth-langs:isp}Ungrounded elements]{\includegraphics[scale=0.4]{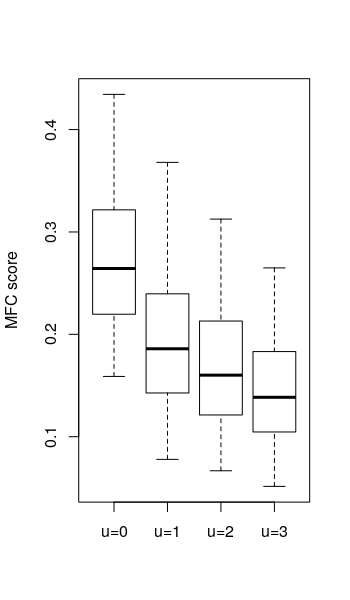}} \quad
    \subfloat[\label{fig:synth-langs:par}Paraphrase]{\includegraphics[scale=0.4]{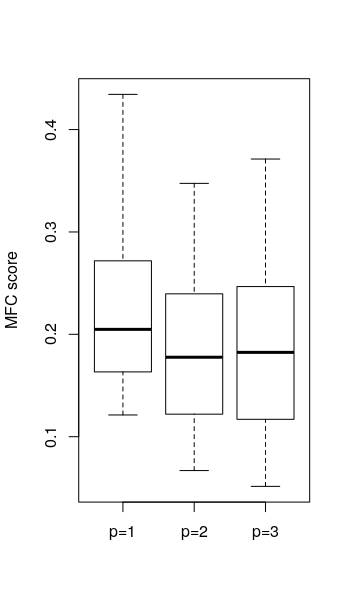}}
    \caption{\textsc{Mfc} for \editsTM{artificial} languages, grouped by parameter. {\smaller (significant items only, avg. of 50 runs)}}
    \label{fig:synth-langs}
\end{figure}
A visualization of the results for the compositional variation factors is shown in Figure~\ref{fig:synth-langs}. Each subfigure corresponds to a different factor, and shows the distribution of \textsc{mfc} scores according to the possible levels for that factor. As expected, random baselines were found to be insignificant (p-value $\geq 0.05$).

If we focus on compositionality (Figure \ref{fig:synth-langs:hol}), we do see that less compositional languages yield lower \textsc{mfc} scores.
When we consider the correlation values averaged over all 50 runs, we see that no holistic parameter configuration (where $h=5$) is found to be significant, resulting in the missing boxplot in Figure \ref{fig:synth-langs:hol}. 
1\textsuperscript{st}, 2\textsuperscript{nd} and 3\textsuperscript{rd} quartiles are found to consistently decrease for higher values of $h$.%
\footnote{
    Some languages with $h=5$, which are fully holistic, were found to yield significant \textsc{mfc}.
    Most of these also included multiple ungrounded symbols ($u \geq 1$).
    This can be explained by the effects of paraphrases: in holistic languages with multiple messages per meaning containing ungrounded symbols, paraphrastic messages for a given meaning differ only by their ungrounded symbols, whereas messages for different meanings will also differ by their grounded symbols---leading to nonzero correlation. 
    However, on average over all 50 runs, the p-value for any of these settings is below our threshold.
}

Synonymy and semantically ungrounded elements are found to be confounding factors (Figures \ref{fig:synth-langs:syn} and \ref{fig:synth-langs:isp}). Higher values for the $s$ and $u$ parameters systematically entail that the distribution of \textsc{mfc} scores (averaged over 50 runs) is globally lower, as  1\textsuperscript{st}, 2\textsuperscript{nd} and 3\textsuperscript{rd} quartiles consistently decrease. 

Lastly, for non fully-holistic settings ($h < 5$), we observe that some combinations of factors fail to produce significant \textsc{mfc} scores.
In $15.2~\%$ of all possible combinations, as this persists even when averaged over all runs, we conclude that this is an actual effect of the interaction of factors.
All these languages are defined with at least one extreme factor: viz. either three synonyms per concept ($s=3$), three ungrounded symbols ($u=3$) or four concepts merged into a single expression ($h=4$); moreover all of them (except for two languages defined with $h=4$, $s=3$ and either $u=2$ or $u=3$) contained a single message per meaning ($p=1$); confirming this trend, we find that all non fully-holistic languages with up to three messages per meaning ($p=3$) were found to have a significant \textsc{mfc} on average.
These shared characteristics can hint at the fact that confounding factors can significantly obfuscate the compositional structure of the messages. An alternative explanation could be that languages without paraphrases ($p=1$) contain fewer messages and thus yield higher p-values, whereas paraphrases additionally entail that very low textual distances map to zero meaning distances. 
%

\subsection{Discussion \& Conclusions}

\begin{table}[t]
    \centering
    \input{tables/parole-sl-lm}
    \caption{\editsTM{Linear model of correlation with 
    parameters as predictors. {\smaller Intercept: $h=1,s=1,u=0,p=1$.}} 
    }
    \label{tab:lm-synthetic-languages}
\end{table}
We observed that synonymy (Figure \ref{fig:synth-langs:syn}) and ungrounded elements (Figure \ref{fig:synth-langs:isp}) seemed detrimental to \textsc{mfc} scores, whereas the effects of paraphrases were found to be more subtle (Figure \ref{fig:synth-langs:par}).
We quantify this by computing a simple linear model \editsTM{in \textsc{r}} \editsTB{\cite{r_core_team_r_2013}} where the correlation score is the dependent variable and the values of the four parameters are the predictors; data points correspond to specific runs. 
Results are reported in Table~\ref{tab:lm-synthetic-languages}. 
While $h=5$ was found to be the predictor with the strongest negative effect on \textsc{mfc} scores, we found that factors $s=3$, $u=3$ and $u=2$ had stronger effects than $h=4$. 
In short, the model shows that factors such as synonymy impact \textsc{mfc} measurements---sometimes to a greater extent than compositionality \editsTM{as shown by t-value scores.
It also stresses that paraphrases positively impact \textsc{mfc} scores}
: in languages where $p>1$, any single unreliable message is less likely to whittle down scores.

In all, our experiment suggests that taking \textsc{mfc} as an indicator for compositionality comes with significant challenges. Factors that we expect from natural language, such as ungrounded symbols and synonyms, obfuscate the clear relationship between compositionality and \textsc{mfc} scores.
At times, these factors can even annihilate \textsc{mfc} scores of compositionally generated languages.
Yet compositionality does impact measurements: therefore, while \textsc{mfc} scores in and of themselves may not be sufficient to establish or reject the compositionality of a given set of messages, they can serve as a diagnosis tool. 


\section{\textsc{Mfc} and \editsTM{N}atural \editsTM{L}anguage: \editsTM{D}efinitions}
\label{sec:defin}

The experiments we conducted in Section~\ref{sec:synthlang} have revealed that confounding factors can impact \textsc{mfc} scores on \editsTM{artificial} languages. This raises the question of how \textsc{mfc} behaves on \emph{natural} instances of compositionality: can the same confounding variables be observed? Is it possible to control for them? In short, we now aim to translate our results derived from \editsTM{artificial} data to natural language data.

Computing Mantel tests for natural examples of composition requires us to select a set of natural language expressions for which we can compute semantic distances between any pair.
Taking inspiration from \newcite{hill-etal-2016-learning-understand}, we use dictionary definitions.
Indeed, in dictionaries, 
the meaning of the word being defined (or ``definiendum''; pl. ``definienda'') is arguably also the meaning of the definition gloss. 
%
Hence, we can equate the semantic distance between two glosses to the semantic distance between the corresponding definienda. 
Distributional semantics models \cite{lenci2018distributional} allow us to conveniently compute the latter: they represent words as vectors in a space over which we can use metrics such as cosine similarity or Euclidean distance, and have \editsTM{moreover} been found to match human intuitions \cite{mandera2017Explaining}.

\subsection{Methodology} 

We select definitions from the dataset distributed by \newcite{noraset_defmod_2017}.
We restrict ourselves to definitions of nouns collected from the \textsc{gcide}, where the definiendum is among the $100$--$10~000$ most frequent words of the English Gutenberg corpus.\footnote{%
    The GNU Collaborative International Dictionary of English (\textsc{gcide}): \url{https://gcide.gnu.org.ua/}.
}
This yields $4~123$ distinct definienda and $20~109$ definitions.
We repeat this process five times for all subsequent measurements before averaging results.

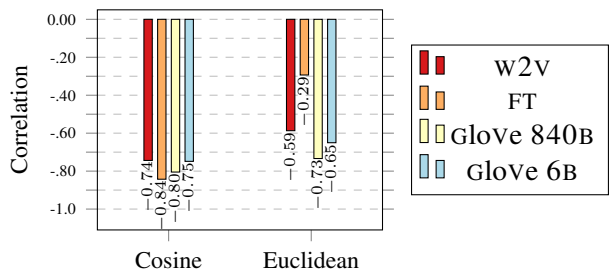
\begin{figure}[t]
    \centering

    \input{figs/tex/figure_meaningdistances_MEN}

    \caption{
        Meaning distance metrics evaluated on the \textsc{men} dataset. \smaller{(Spearman correlation)}
    }
    \label{fig:MEN}
\end{figure}

One could arguably compute \textsc{mfc} on the basis of human-annotated semantic similarity ratings.
Datasets of such ratings exist \cite[e.g.]{SICK,cer-etal-2017-semeval}, but they do not provide a (dis)similarity between every pair of items as is required to perform a Mantel test.
Instead, as mentioned above, we use distributional semantics models as proxy for meaning representations and use a distance between word vectors.
We consider four sets of pre-trained word embeddings:
{fastText} 
\cite{bojanowski2016enriching}, {\textsc{g}lo\textsc{v}e 6\textsc{b}} 
and {\textsc{g}lo\textsc{v}e 840\textsc{b}} 
\cite{pennington-etal-2014-glove},
and {word2vec} 
\cite{mikolov13efficient}.

We first verify that the semantic distances over these semantic spaces properly anti-correlate to human similarity ratings: words that humans judge to be highly similar in meaning should not be far from one another in the embedding spaces. 
We therefore assess the four sets of word embeddings on a word semantic similarity benchmark, the \textsc{men} dataset \cite{MEN}; for all models, we test the Euclidean distance and the cosine distance as a distance over vectors. 
\editsTM{Results in Figure~\ref{fig:MEN} highlight that while all semantic metrics properly anti-correlate, 
cosine distance yields higher correlations with human similarity judgments than Euclidean distance, for all embedding spaces.}

While it is most standard to build on cosine (the most common semantic vector similarity metric) and Levenshtein distance (consistently with artificial language studies), we also consider alternative setups with \editsTM{Euclidean 
distance, tree-based form metrics and distance normalization. }
Unlike the \editsTM{artificial languages we devised earlier, }
natural language has 
a rich syntax.
It therefore makes sense to 
assess the textual similarity of these sentences using syntactically-informed metrics.
Thus, in addition to the Levenshtein distance over words, we study the Tree Edit Distance (\textsc{ted}) computed with the \textsc{ap-ted} algorithm \cite{pawlik15apted} over the corresponding parse trees obtained with the Reconciled Span Parser \cite{joshi-etal-2018-extending}.
As textual distances should arguably not be sensitive to the size of the sentences, we also consider normalizing them so that the maximum distance between any two definitions 
is $1$.\footnote{%
    We normalize Levenshtein distance by the length of the longest of the two items.
    We normalize \textsc{ted} between $\mathcal{F}$ and $\mathcal{G}$ by $\#\mathcal{F} + \#\mathcal{G} - \min\big(\textit{h}(\mathcal{F}), \textit{h}(\mathcal{G})\big)$, where $\#\mathcal{T}$ corresponds to the size of tree $\mathcal{T}$ and $\textit{h}(\mathcal{T})$ to its height.
} 

To check whether synonymy, ungrounded symbols and paraphrases also impact natural language examples, we perform simple modifications of our original process. 
To control for synonymy, we replace every word by the first lemma of its first synset in WordNet \cite{fellbaum98wordnet}, if any such lemma can be found. 
%
To control for ungrounded symbols, we remove stop-words from our definitions; as this results in definitions that greatly differ from the parser's training examples, we do not compute \textsc{ted}-based \textsc{mfc} scores in this case.
To control for paraphrases, we redo the selection process, and this time randomly sample $4~123$ items out of the total $20~109$ definitions, without the constraint of having only one definition per definiendum: hence these samples only contain on average $2~507$ distinct definienda (standard deviation: $\pm 14.53$). 
\editsTM{We duly note that different definitions correspond to distinct senses of the definiendum and thus are not strictly speaking paraphrases. However, \newcite{Arora18LinearWordsenses} show that the embedding of a polysemous word corresponds to a weighted sum of underlying meaning embeddings; thus a word embedding ought to be matched to the entire set of definitions for the corresponding token.}
%
\editsTM{In any event,} we expect these methods to indicate a consistent trend in \textsc{mfc} scores\editsTM{, despite their crudeness.}

\subsection{Results} 

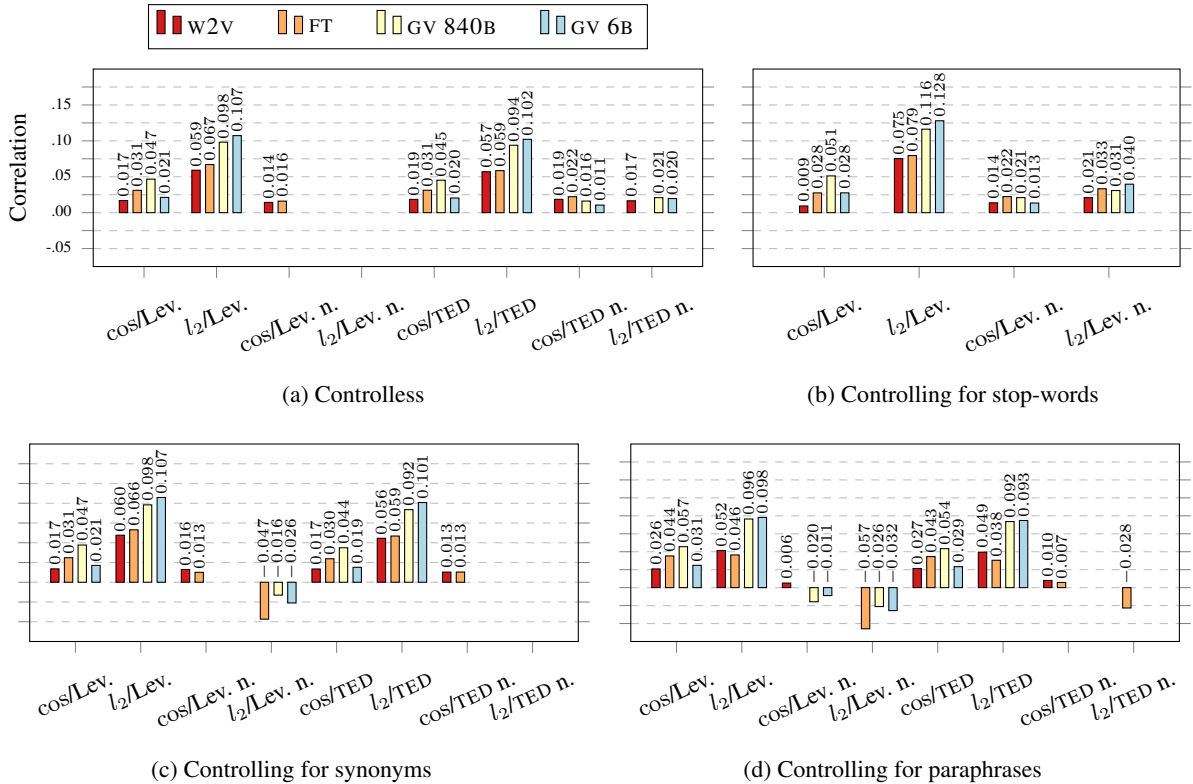
\begin{figure}[t]
    \centering

    \input{figs/tex/figure_defs-all}

    \caption{
        \textsc{Mfc} scores for natural language definitions.
    }
    \label{fig:mfc-def}
\end{figure}

We summarize our findings in Figure~\ref{fig:mfc-def}.
Figure~\ref{fig:mfc-def:no-ctrl} corresponds to \editsTB{results in the case where no supplementary control is applied.} 
Figures~\ref{fig:mfc-def:stop-words}, \ref{fig:mfc-def:syn} and \ref{fig:mfc-def:par} highlight the effects of controlling for stop-words, synonyms, and multiple definitions \editsTM{respectively}. Statistically insignificant \editsTB{\textsc{mfc} scores} 
are not displayed. 
\editsTM{Standard deviations of \textsc{mfc} scores across the five runs remain below $0.008$ and often around $0.005$, with the exception of a few setups when controlling for paraphrases.\footnote{
    \editsTM{In detail, 4 setups yield a standard deviation between $0.01$ and $0.015$; they all involve Euclidean distance and non-normalized form distances, using either fastText or \textsc{g}lo\textsc{v}e 840\textsc{b}.}
}}

%
%
%
While our \editsTM{crude control methods 
often produce inconsistent measures}, they do suggest that the identified \textsc{mfc} confounds are noticeably at play in natural language. 

As with artificial language experiments above, controlling for \editsTM{\emph{stop-words}} improved \textsc{mfc} measurements in most setups (Figure~\ref{fig:mfc-def:stop-words})\editsTM{, yet} scores decrease for fastText and word2vec when using cosine and Levenshtein distances setups.
%
Controlling for \editsTM{\emph{synonymy}} brought very small suggestive \textsc{mfc} increases in the most standard setup (cosine and non-normalized Levenshtein) but observations across other setups are not consistent (Figure~\ref{fig:mfc-def:syn}).
%
Controlling for \editsTM{\emph{paraphrases}} produces consistent and pronounced \textsc{mfc} improvements on the most standard setup but quite diverse effects elsewhere (Figure \ref{fig:mfc-def:par}).

\editsTM{We also} make two general unexpected observations about alternative setups.
First, as seen e.g., from the controlless scenario in Figure~\ref{fig:mfc-def:no-ctrl}, normalizing textual metrics can be surprisingly detrimental.
For instance, Euclidean distance between {\textsc{g}lo\textsc{v}e 6\textsc{b}} vectors, when paired with Levenshtein distance, goes from the highest measured \textsc{mfc} score to statistical insignificance---in fact, only $2$ out of the $8$ \textsc{mfc} using normalized Levenshtein distance are significant. When controlling for \editsTM{confounding variables}, normalization can even induce anti-correlations.
\editsTM{Second, cosine distance also yields 
}lower \textsc{mfc} scores than Euclidean distance, despite it being more in line with human ratings (Figure~\ref{fig:MEN}):
Euclidean distance yields in many occasions correlation scores of $0.1$ and higher, more than twice what we observe for cosine distance\editsTM{.
}

\subsection{Discussion \& Conclusions}

Overall, this experiment highlights that demonstrating the role of confounding factors that we expect to be detrimental to \textsc{mfc}, such as ungrounded elements, is in principle possible for natural language; 
though we have employed blunt methods of control, effects could be perceived.
On the other hand, our observations underscore how sensitive \textsc{mfc} is to the choice of distance functions: considerations such as normalizing metrics between $0$ and $1$ or choosing Euclidean vs.\ cosine distance can impact results significantly. 
These observations raise two questions: (i) why does normalizing textual distances degrade \textsc{mfc} scores? and (ii) do these results support that \textsc{mfc} captures compositionality in natural language? To answer both, we study more closely which items are detrimental to \textsc{mfc}, as they may shed light on what \textsc{mfc} measurements capture, and how normalization affects it. We consider items where measurements are mismatched: sentences with a relatively low meaning distance but a relatively high form distance---or vice versa---drive the \textsc{mfc} score down.
\editsTM{W}e convert distance measurements into rank values, and consider the $100$ \editsTB{pairs of sentences that yield maximal rank difference for a given setup.
These pairs will be referred to as \emph{problematic} below.}

For the sake of clarity, we focus on \textsc{g}lo\textsc{v}e 6\textsc{b}-based measurements, using Levenshtein distance (either raw or normalized) and compute all our measurements from the same random selection of $4~123$ definitions.
In the controlless scenario, the $100$ most problematic items for non-normalized Levenshtein-based setups all involved two synonym-based definitions (e.g., \exInText{pilot: a steersman}).
As synonym-based definitions can be identified to holistic messages, this evaluation suggests that our dataset contains a high number of non-compositional 
examples that this textual metric is not fit to handle. 
That these items are reliably deemed problematic is coherent with the hypothesis that \textsc{mfc} captures compositionality. 

All problematic pairs were found to have a high semantic distance (involving two unrelated definienda) but a low textual distance (both definitions are very short, usually composed of an article and a noun, which entails few edit operations).
This suggests that normalizing the textual distance might be crucial to get reliable \textsc{mfc}.
However, while normalizing Levenshtein distance does reduce the number of pairs of synonymy-based definitions in the problematic pairs, they remain very frequent ($98/100$ for Euclidean distance, $88/100$ for cosine distance), due to the fact that a common article usually means that half of the tokens in the two definitions are the same.
Removing stop-words further reduces synonymy-based definitions in problematic pairs to a handful, but reveals other artifacts: many pairs share a common pattern such as 
\exInText{sneer: the act of sneering}/\exInText{wade: the act of wading}.
Moreover, the \textsc{mcf} obtained in these settings are lower than with raw Levenshtein distance, suggesting that such patterns are frequent and that length is (counter-intuitively) a relevant factor.
We conjecture that short definitions are responsible for a large part of these artifacts, in which case a solution would be to filter them out.
Alternatively, a more complex textual distance could allow us to get rid of these artifacts.
We leave this subject for future work.

In summary, this experiment suggests that \textsc{mfc} can detect compositionality in natural language, but that the exact setup employed is crucial.
Distance normalization appears to annihilate \textsc{mfc} scores and the effects of controlling for confounding factors vary from distances to distances.


\section{\textsc{Mfc} and Natural Language: Sentence Encoders}
\label{sec:sentenc}

The results in Section \ref{sec:defin} suggest that \textsc{mfc} scores on natural language reflect its compositionality but also confounding structural factors that obfuscate semantic transparency.  However, the findings also raise questions about the peculiarity of the dataset, since dictionary definitions do not form a representative sample of language use by any standards. Is it possible to replicate the methodology on more naturalistic text data?
The main difficulty lies in finding a relevant and practical notion of semantic distance for arbitrary sentences, since human judgments can prove arduous to collect.
As an attractive potential strategy, one could use vector distances over semantic representations produced by models of semantic composition called sentence encoders. Do these models provide good enough approximations for sentence meaning similarity? If so, we expect to be able to reproduce the effects of confounding factors on \textsc{mfc} as predicted by representations derived from sentence encoders.
To answer these questions, we replicate our previous experiment on a set of common sentences, rather than definitions.

\subsection{Methodology}

The semantic distances used in this section rely on
\textit{sentence encoders}, computational models that convert sequences of tokens into vector representations.
They can be trained on a variety of tasks, from predicting the entailment relation between a pair of sentences \cite{conneau-EtAl:2017:EMNLP2017} to reconstructing the context of a passage \cite{kiros2015skipthought}.
These tasks require capturing the meaning of the corresponding texts.

\begin{figure}[t]
    \centering
    \input{figs/tex/figure_meaningdistances_SICK}
 
    \caption{%
        Meaning distance metrics evaluated on the \textsc{sick} dataset. {\smaller (Spearman correlation)}
    }
    \label{fig:SICK}
\end{figure}
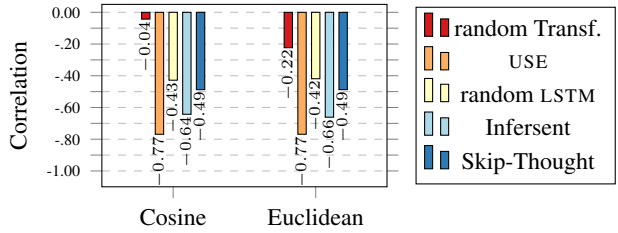

We first assess whether sentence encoders yield coherent representations of meaning by computing the Spearman correlation between the human ratings present in the \textsc{sick} benchmark \cite{SICK} and the cosine and Euclidean distances between the two corresponding sentence embeddings; as with word embedding spaces in Section~\ref{sec:defin}, we expect a significant anti-correlation between the two.
Figure~\ref{fig:SICK} summarizes correlation scores for Skip-Thought \cite{kiros2015skipthought}, Infersent \cite{conneau-EtAl:2017:EMNLP2017} and the Universal Sentence Encoder \cite[\textsc{use} for short]{cer-etal-2018-universal}, along with a randomly initialized (and untrained) Transformer \cite{vaswani_attention_2017} for perspective. 
We observe that \textsc{use} yields the most consistent semantic representations and thus decide to focus in the following on this particular model.

We randomly sample $4~123$ sentences from the Toronto BookCorpus \cite{bookcorpus}---a collection of English books of various genres---for computing Mantel tests using Levenshtein distance, both raw and normalized, as the textual distance.
We repeat the procedure $5$ times before averaging results.
We use the same operations as previously to control for synonyms and stop-words; i.e., we test whether removing stop-words altogether and normalizing words based on their WordNet synsets improve \textsc{mfc}. 

\subsection{Results}

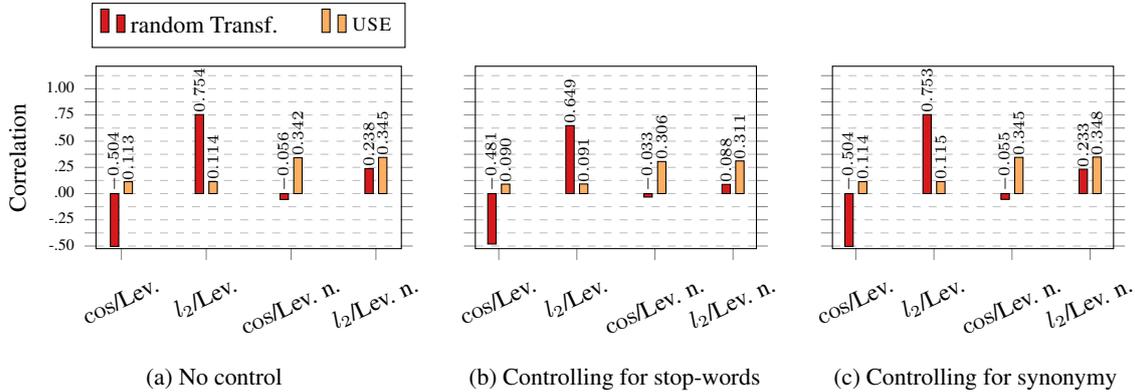
\begin{figure}[t]
    \centering
    \input{figs/tex/figure_sents-all}
 
    \caption{%
        \textsc{Mfc} scores for natural language sentences.
    }
    \label{fig:mfc-sent}
\end{figure}

Results are presented in Figure~\ref{fig:mfc-sent}: Figure~\ref{fig:mfc-sent:no-ctrl} corresponds to the controlless scenario; Figures \ref{fig:mfc-sent:isp} and \ref{fig:mfc-sent:syn} present the effects of controlling for stop-words and synonyms respectively. \editsTM{Results are consistent across all five random samples of sentences: standard deviation of \textsc{mfc} scores is systematically below $0.016$, and often below $0.005$.\footnote{
    \editsTM{Only the randomly initialized Transformer when using Euclidean distance and non-normalized Levenshtein distance yields standard deviations above $0.01$ (in all three scenarios). Standard deviations for \textsc{use} embeddings are all below $0.003$, with the exception of the two setups involving normalized Levenshtein distance and controlling for stop-words (around $0.007$).}
}}

Most striking are the very high correlations and anti-correlations that are yielded by the random baseline: the anti-correlations and correlations derived from non-normalized Levenshtein distance have a greater magnitude than what we observe for \textsc{use} (which is also based on the Transformer architecture).
In the case of Euclidean distance, this magnitude can partly be explained by the architecture itself, which computes a vector that is not meaningful but that is still computed in a very compositional way, although the corresponding notion of composition is not linguistically justified.
The simple sum used to derive the sentence embeddings from the hidden states at each timestep\footnote{%
    Like \newcite{cer-etal-2018-universal}, we divided the sentence embedding by the square root of the length of sentence.
} entails that the norm of every sentence representation grows proportionally to the number of words it contains; moreover the residual connections used in Transformers entail that the hidden state for a given timestep bears some trace of the input word at this timestep, thus sentences with words in common will tend to be nearer in the Euclidean space.

Turning to the \textsc{mfc} for \textsc{use}, we observe that normalizing Levenshtein distance leads to higher scores---opposite to what we found for definitions\editsTM{.} 
\editsTM{On the other hand, cosine-based setups overall are found to decrease scores 
by a low margin, of $0.005$ at most---which is consistent with what was observed for definitions, but much less pronounced.} 
Removing stop-words lowers the correlation for \textsc{use} embeddings, which  \editsTM{is also contrary to} what we observed for definitions:
scores in Figure \ref{fig:mfc-sent:isp} are found to be lower 
\editsTB{than those without any form of control} 
by a margin ranging from $0.02$ to $0.04$.
\editsTB{Lastly, while}
we technically observe a higher \textsc{mfc} when controlling for synonyms (Figure \ref{fig:mfc-sent:syn}), the effect is again very subtle: correlation \editsTM{increases by no} more than $0.005$\editsTM{.} 
\subsection{Discussion \& Conclusions}

This last experiment is first a reminder of the fact that the use of sentence encoders to compute \textsc{mfc} only makes sense under the assumption that they accurately represent the semantic of their input.
While this assumption is obviously not satisfied for a random model, it is defensible for \textsc{use} given its training procedure and the high anti-correlation observed above in Figure~\ref{fig:SICK}.

Second, if we accept the relevance of using this latter model, the results for \textsc{use} highlight the fact that the issues we encountered with the definition dataset of Section~\ref{sec:defin}, in particular related to distance normalization, are not systematically present.
This suggests that the counter-intuitive behavior observed on definitions might be due to particular artifacts from this dataset, \editsTM{in contrast to the more natural and varied dataset used in the present section.}


\section{Related \editsTM{W}orks}

The core of this work draws upon previous research on \textsc{mfc} \cite[a.o.]{kirby08cumul,Kirby2015CompressionAC,spike2017minimal-thesis,Ren2020Compositional}.
This concept finds its roots in works such as the ones of \newcite{kirby99syntaxOutOfLearning}, \newcite{Kirby2001SpontaneousEvolution} or \newcite{brighton2006Topo}, which suggest that while compositionality in language relies on syntactic structures, it could still be measured using surface form. 
Other implementations of \textsc{mfc} include studies centered on correlations between form and meaning at the word or sub-morphemic level---conflicting with the assumption of arbitrariness of the sign---such as the work of \newcite{gutierrez-etal-2016-finding}, \newcite{Kutuzov2017ArbitrarinessOL}, \newcite{dautriche17wfsim} or \newcite{pimentel-etal-2019-meaning}.

More broadly, compositionality has proven to be a fruitful field of research; see for example the overview edited by \newcite{TheOxfordHandbookofCompositionality}.
The \textsc{nlp} community has produced models of compositional semantics \cite{Baroni2014FregeIS,Marelli2015AffixationIS,conneau-EtAl:2017:EMNLP2017} and significant efforts have been made to assess how well they capture human intuitions \cite[a.o.]{dinu-baroni-2014-make,SICK,cer-etal-2017-semeval,wang-etal-2018-glue} and to analyze their behavior \cite[e.g.]{Liska18Haystack,Baroni20LinguisticGen}. 
\editsTM{It is worth stressing that assessing the degree of compositionality within a corpus or a model, as do these works, is a distinct enterprise from quantifying the degree of similarity between two sentences \cite[e.g.]{papineni-etal-2002-bleu,lin-2004-rouge,clark-etal-2019-sentence}---more precisely, the methodology of \textsc{mfc} employs the latter to estimate the former.}

Lastly, another related topic of research concerns the use of dictionaries as meaning inventories in \textsc{nlp}.
For instance, \newcite{chodorow-etal-1985-extracting}, \newcite{Gaume2004WSDict}, \newcite{tissier-etal-2017-dict2vec} and \newcite{bosc-vincent-2018-auto}, all employ dictionary entries for purposes ranging from word-sense disambiguation to ontology population and to embedding computation. 
Directly relevant here is the work of \newcite{hill-etal-2016-learning-understand}, who leverage the compositional aspect of definitions to compute sentence meaning representations. 

\section{Conclusions}

In this work, we have empirically assessed the validity of meaning-form correlation as a methodology for studying compositionality, both in artificial and natural languages. 
In all, the relationship between meaning-form correlation and compositionality is not straightforward; experimental setups that would employ the former to represent the latter ought to proceed with caution.
Confounding factors that we can expect from natural language can also weigh on meaning-form correlation in \editsTM{artificial} languages, and their effects are not straightforward to control for.
Other issues such as the architecture employed to derive meaning representations weigh on the relevance or interpretation of correlation scores.
Meaning-form correlation is very dependent on the actual metrics and dataset under consideration. 
On the other hand, meaning-form correlation was also shown to be able to detect compositionality, both in controlled \editsTM{artificial} setups and with natural language data, suggesting that this methodology can be used to study compositionality as long as due attention is paid to confounding factors.

While we have focused on modeling a small number of confounding factors, others can obviously be tested using our methodology.
For instance, preliminary experiments suggest that factors such as duality of patterning---i.e., using non-atomic expressions \cite{Hockett1960OrigSpeech}---reinforce correlations to some extent, whereas aspects such as free ordering of constituents seem detrimental. 
This limited study has also focused on a specific implementation of meaning-form correlation, based on Mantel tests---however other related setups exist, for example based on mutual information \cite{pimentel-etal-2019-meaning}. 
We leave them to future studies.
Another limitation of the present work is that we have focused solely on English: we intend in future experiments to experiment with typologically diverse languages\editsTM{, ranging the full isolating-synthetic spectrum}.

\section*{\editsTB{Acknowledgements}}

\editsTM{We thank Hiroya Takamura for discussions on this work.} 

\editsTB{This paper is based on results obtained from a project commissioned by AIST-Tokyo Tech Real World Big-Data Computation Open Innovation Laboratory (RWBC-OIL).} 
\editsTM{The work was also supported  by a public grant overseen by the French National Research Agency (ANR) as part of the ``Investissements d'Avenir'' program: Idex \emph{Lorraine Universit\'e d'Excellence} (reference: ANR-15-IDEX-0004).} 

\bibliographystyle{coling}
\bibliography{coling2020}

\end{document}

%% file: tables/parole-sl-lm.tex
\footnotesize\npdecimalsign{.}
\begin{tabular}{l n{2}{6} n{1}{6} n{4}{2} l}
\hline
\textbf{\small Coeffs.}&\textbf{\small Estimate}&\textbf{\small Std. Error}&\textbf{\small t value}&\textbf{\small Pr($>|t|$)}\\\hline
\textit{\smaller(Intercept)} & 0.402781 & 0.001936 & 208.07 & $< 2 \cdot 10^{-16}$\\\hline
$h=2$ & -0.016909 & 0.001577 & -10.72 & $< 2 \cdot 10^{-16}$\\
$h=3$ & -0.057501 & 0.001577 & -36.46 & $< 2 \cdot 10^{-16}$\\
$h=4$ & -0.113547 & 0.001577 & -72.00 & $< 2 \cdot 10^{-16}$\\
$h=5$ & -0.197191 & 0.001734 & -113.75 & $< 2 \cdot 10^{-16}$\\\hline
$s=2$ & -0.096726 & 0.001748 & -55.34 & $< 2 \cdot 10^{-16}$\\
$s=3$ & -0.137808 & 0.001748 & -78.85 & $< 2 \cdot 10^{-16}$\\\hline
$u=1$ & -0.099435 & 0.001524 & -65.26 & $< 2 \cdot 10^{-16}$\\
$u=2$ & -0.126905 & 0.001524 & -83.29 & $< 2 \cdot 10^{-16}$\\
$u=3$ & -0.145707 & 0.001524 & -95.63 & $< 2 \cdot 10^{-16}$\\\hline
$p=2$ & 0.031989 & 0.001354 & 23.63 & $< 2 \cdot 10^{-16}$\\
$p=3$ & 0.041219 & 0.001354 & 30.45 & $< 2 \cdot 10^{-16}$\\\hline
\end{tabular}

%% file: figs/tex/figure_meaningdistances_MEN.tex
\begin{tikzpicture}
\begin{axis}[
    ybar,
    width=\textwidth/3,
    height=4.5cm,
    bar width=.11cm,
    legend style={at={(1.1,0.5)},
      anchor=west,legend columns=1,
      /tikz/every even column/.append style={column sep=0.5cm}},
    ylabel={\footnotesize Correlation},
    ytick={-1.0,-0.9,..., 0.1},
    yticklabels={-1.0, ,-.80, ,-.60, ,-.40, ,-.20, ,0.00},
    ymin=-1.11, ymax=0.05,
    ytick align=outside,
    ymajorgrids=true,
    grid style=dashed,
    symbolic x coords={cdist,l2,},%
    xtick=data, xtick pos=bottom,
    enlarge x limits=0.5,
    xticklabels={Cosine,Euclidean},
    ticklabel style = {font=\small},
    y tick label style = {font=\tiny},
    nodes near coords,
    nodes near coords style={
        font=\tiny,
        rotate=90,
        anchor=east,
        xshift=0.3125em,
        /pgf/number format/fixed,
        /pgf/number format/zerofill,
        /pgf/number format/precision=2,
        color=black,
    },
    cycle list/RdYlBu-5,
    every axis plot/.append style={
        fill,
    },
    ]

    \addplot+[draw=black,] coordinates {
     (cdist, -0.7437986070621162)
     (l2,  -0.5869993182447595)
    };
    
    \addplot+[draw=black,] coordinates {
      (cdist,  -0.8426990296251209)
      (l2,  -0.2934388990290073)
    };
    
    \addplot+[draw=black,] coordinates {
     (cdist, -0.804934472292358)
     (l2,  -0.7336334027622713)
    };
    
    \addplot+[draw=black,] coordinates {
     (cdist, -0.7485685413592661)
     (l2, -0.6503385177561019)
    };

\legend{\textsc{w2v}, \textsc{ft}, \textsc{g}lo\textsc{v}e \textsc{840b}, \textsc{g}lo\textsc{v}e \textsc{6b}};

\end{axis}
\end{tikzpicture}

%% file: figs/tex/figure_defs-all.tex
\subfloat[\label{fig:mfc-def:no-ctrl}Controlless]{
\begin{tikzpicture}
\begin{axis}[
    ybar,
    width=0.6\textwidth,
    height=4.2cm,
    bar width=.11cm,
    ylabel={\footnotesize Correlation},
    ytick={-0.05 ,-0.025, 0.0, 0.025, 0.05, 0.075, 0.1, 0.125, 0.15, 0.175},
    yticklabels={-.05, , .00, , .05, , .10, ,.15,},
    ymin=-0.075, ymax=0.2,
    ytick align=outside,
    ymajorgrids=true,
    grid style=dashed,
    symbolic x coords={cdistlevenshtein,l2levenshtein,cdistlevenshteinn,l2levenshteinn,cdistapted,l2apted,cdistaptedn,l2aptedn},
    xtick=data, xtick pos=bottom,
    xticklabels={
        $\cos$/Lev., $l_2$/Lev., $\cos$/Lev. n., $l_2$/Lev. n., 
        $\cos$/\textsc{ted}, $l_2$/\textsc{ted}, $\cos$/\textsc{ted} n., $l_2$/\textsc{ted} n.,
        },
        ticklabel style = {font=\footnotesize},
        x tick label style = {rotate=25},
        y tick label style = {font=\tiny},
        typeset ticklabels with strut,
        major y tick style={
            /pgfplots/major tick length=1.5mm,
        },
        nodes near coords always on top/.style={
            scatter/position=absolute,
            positive value/.style={
                at={(axis cs:\pgfkeysvalueof{/data point/x},\pgfkeysvalueof{/data point/y})},
            },
            negative value/.style={
                at={(axis cs:\pgfkeysvalueof{/data point/x},0)},
            },
            every node near coord/.append style={
                check values/.code={%
                    \begingroup
                    \pgfkeys{/pgf/fpu}%
                    \pgfmathparse{\pgfplotspointmeta<0}%
                    \global\let\result=\pgfmathresult
                    \endgroup
                    %
                    %
                    \pgfmathfloatcreate{1}{1.0}{0}%
                    \let\ONE=\pgfmathresult
                    \ifx\result\ONE
                        \pgfkeysalso{/pgfplots/negative value}%
                    \else
                        \pgfkeysalso{/pgfplots/positive value}%
                    \fi
                },
                check values,
                anchor=west,
                rotate=90,
                font=\tiny,
                /pgf/number format/fixed,
                /pgf/number format/zerofill,
                /pgf/number format/precision=3,
                xshift=-0.5ex,
                color=black,
            },
        },
        nodes near coords={
            \pgfmathprintnumber[fixed zerofill,precision=3]{\pgfplotspointmeta}
        },
        nodes near coords always on top,
        legend style={at={(0.5,1.1)},
            anchor=south,legend columns=-1,
            /tikz/every even column/.append style={column sep=0.5cm}
        },
        cycle list/RdYlBu-5,
        every axis plot/.append style={
            fill,
        },
    ]

\addplot [white,only marks,forget plot, nodes near coords={}] coordinates {
  (cdistlevenshtein,.13)
  (l2levenshtein,.13)
  (cdistlevenshteinn,.13)
  (l2levenshteinn,.13)  
  (cdistapted,.13)
  (l2apted,.13) 
  (cdistaptedn,.13)
  (l2aptedn,.13) 
}; 

\addplot+[draw=black,error bars/.cd,y dir=both,y explicit relative,] coordinates {
(cdistlevenshtein, 0.016906309115858928) 
(l2levenshtein, 0.05913940568985963) 
(cdistlevenshteinn, 0.014297867010866282) 
(cdistapted, 0.018525446527811518) 
(l2apted, 0.05697385916623469) 
(cdistaptedn,0.018585) 
(l2aptedn,0.016530) 

}; 

\addplot+[draw=black,error bars/.cd,y dir=both,y explicit relative,] coordinates {
(cdistlevenshtein, 0.03110060198704507) 
(l2levenshtein, 0.06701253742369366) 
(cdistlevenshteinn, 0.016089142174786162) 
(cdistapted, 0.03125705081863811) 
(l2apted, 0.058527225519962534) 
(cdistaptedn,0.022265) 

}; 

\addplot+[draw=black,error bars/.cd,y dir=both,y explicit relative,] coordinates {

(cdistlevenshtein, 0.046696391099520174) 
(l2levenshtein, 0.09812904627722754) 
(cdistapted, 0.04522271905292958) 
(l2apted, 0.09391747704563379) 
(cdistaptedn,0.016175) 
(l2aptedn,0.020944) 

}; 

\addplot+[draw=black,error bars/.cd,y dir=both,y explicit relative,] coordinates {

(cdistlevenshtein, 0.021265281323303752) 
(l2levenshtein, 0.10730223626683119) 
(cdistapted, 0.020429245831540432) 
(l2apted, 0.10217748824958664) 
(cdistaptedn,0.010570) 
(l2aptedn,0.019661) 

}; 


\legend{{\footnotesize\textsc{w2v}}, {\footnotesize\textsc{ft}}, {\footnotesize{\textsc{gv 840b}}}, {\footnotesize{\textsc{gv 6b}}}};

\end{axis}
\end{tikzpicture}}
\subfloat[\label{fig:mfc-def:stop-words}Controlling for stop-words]{
\begin{tikzpicture}
\begin{axis}[
    ybar,
    width=0.45\textwidth,
    height=4.2cm,
    bar width=.11cm,
    ytick={-0.05 ,-0.025, 0.0, 0.025, 0.05, 0.075, 0.1, 0.125, 0.15, 0.175},
    yticklabels={},
    ymin=-0.075, ymax=0.2,
    ytick align=outside,
    ymajorgrids=true,
    enlarge x limits=0.25,
    grid style=dashed,
    symbolic x coords={cdistlevenshteinf,l2levenshteinf,cdistlevenshteinfn,l2levenshteinfn},
    xtick=data, xtick pos=bottom,
    xticklabels={
        $\cos$/Lev.,$l_2$/Lev.,$\cos$/Lev. n.,$l_2$/Lev. n.
        },
    ticklabel style = {font=\footnotesize},
        x tick label style = {rotate=25},
        y tick label style = {font=\tiny},
        typeset ticklabels with strut,
        major y tick style={
            /pgfplots/major tick length=1.5mm,
        },
        nodes near coords always on top/.style={
            scatter/position=absolute,
            positive value/.style={
                at={(axis cs:\pgfkeysvalueof{/data point/x},\pgfkeysvalueof{/data point/y})},
            },
            negative value/.style={
                at={(axis cs:\pgfkeysvalueof{/data point/x},0)},
            },
            every node near coord/.append style={
                check values/.code={%
                    \begingroup
                    \pgfkeys{/pgf/fpu}%
                    \pgfmathparse{\pgfplotspointmeta<0}%
                    \global\let\result=\pgfmathresult
                    \endgroup
                    %
                    %
                    \pgfmathfloatcreate{1}{1.0}{0}%
                    \let\ONE=\pgfmathresult
                    \ifx\result\ONE
                        \pgfkeysalso{/pgfplots/negative value}%
                    \else
                        \pgfkeysalso{/pgfplots/positive value}%
                    \fi
                },
                check values,
                anchor=west,
                rotate=90,
                font=\tiny,
                /pgf/number format/fixed,
                /pgf/number format/zerofill,
                /pgf/number format/precision=3,
                xshift=-0.5ex,
                color=black,
            },
        },
        nodes near coords={
            \pgfmathprintnumber[fixed zerofill,precision=3]{\pgfplotspointmeta}
        },
        nodes near coords always on top,
        cycle list/RdYlBu-5,
        every axis plot/.append style={
            fill,
        },
    ]

\addplot [white,only marks,forget plot, nodes near coords={}] coordinates {
  (cdistlevenshteinf,.13) %
  (l2levenshteinf,.13) %
  (cdistlevenshteinfn,.13) 
  (l2levenshteinfn,.13) 
}; 

\addplot+[draw=black,] plot[error bars/.cd,y dir=both,y explicit relative,] coordinates {
(cdistlevenshteinf, 0.009326709420716091) 
(l2levenshteinf, 0.07527598603030687) 
(cdistlevenshteinfn, 0.013724686555119844) 
(l2levenshteinfn, 0.02097942442192002) 

}; 

\addplot+[draw=black,] plot[error bars/.cd,y dir=both,y explicit relative,] coordinates {
(cdistlevenshteinf, 0.02753648640100917) 
(l2levenshteinf, 0.07946352413038264) 
(cdistlevenshteinfn, 0.02238587400533237) 
(l2levenshteinfn, 0.03331326705256908) 

}; 

\addplot+[draw=black,] plot[error bars/.cd,y dir=both,y explicit relative,] coordinates {
(cdistlevenshteinf, 0.05121040391558046) 
(l2levenshteinf, 0.11627775947806622) 
(cdistlevenshteinfn, 0.020981382299674584) 
(l2levenshteinfn, 0.030816412519866332) 

}; 

\addplot+[draw=black,] plot[error bars/.cd,y dir=both,y explicit relative,] coordinates {
(cdistlevenshteinf, 0.027600495179112627) 
(l2levenshteinf, 0.12796721318285326) 
(cdistlevenshteinfn, 0.013432678641566349) 
(l2levenshteinfn, 0.039665352963361274) 

}; 

\end{axis}
\end{tikzpicture}}

\subfloat[\label{fig:mfc-def:syn}Controlling for synonyms]{
\begin{tikzpicture}
\begin{axis}[
    ybar,
    width=0.55\textwidth,
    height=4.2cm,
    bar width=.11cm,
    ytick={-0.05 ,-0.025, 0.0, 0.025, 0.05, 0.075, 0.1, 0.125, 0.15},
    yticklabels={},
    ymin=-0.075, ymax=0.175,
    ytick align=outside,
    ymajorgrids=true,
    grid style=dashed,
    symbolic x coords={cdistlevenshtein,l2levenshtein,cdistlevenshteinn,l2levenshteinn,cdistapted,l2apted,cdistaptedn,l2aptedn},
    xtick=data, xtick pos=bottom,
    xticklabels={
        $\cos$/Lev., $l_2$/Lev., $\cos$/Lev. n., $l_2$/Lev. n., 
        $\cos$/\textsc{ted}, $l_2$/\textsc{ted}, $\cos$/\textsc{ted} n., $l_2$/\textsc{ted} n.,
        },
    ticklabel style = {font=\footnotesize},
        x tick label style = {rotate=25},
        y tick label style = {font=\tiny},
        typeset ticklabels with strut,
        major y tick style={
            /pgfplots/major tick length=1.5mm,
        },
        nodes near coords always on top/.style={
            scatter/position=absolute,
            positive value/.style={
                at={(axis cs:\pgfkeysvalueof{/data point/x},\pgfkeysvalueof{/data point/y})},
            },
            negative value/.style={
                at={(axis cs:\pgfkeysvalueof{/data point/x},0)},
            },
            every node near coord/.append style={
                check values/.code={%
                    \begingroup
                    \pgfkeys{/pgf/fpu}%
                    \pgfmathparse{\pgfplotspointmeta<0}%
                    \global\let\result=\pgfmathresult
                    \endgroup
                    %
                    %
                    \pgfmathfloatcreate{1}{1.0}{0}%
                    \let\ONE=\pgfmathresult
                    \ifx\result\ONE
                        \pgfkeysalso{/pgfplots/negative value}%
                    \else
                        \pgfkeysalso{/pgfplots/positive value}%
                    \fi
                },
                check values,
                anchor=west,
                rotate=90,
                font=\tiny,
                /pgf/number format/fixed,
                /pgf/number format/zerofill,
                /pgf/number format/precision=3,
                xshift=-0.5ex,
                color=black,
            },
        },
        nodes near coords={
            \pgfmathprintnumber[fixed zerofill,precision=3]{\pgfplotspointmeta}
        },
        nodes near coords always on top,
        cycle list/RdYlBu-5,
        every axis plot/.append style={
            fill,
        },
    ]

\addplot [white,only marks,forget plot, nodes near coords={}] coordinates {
  (cdistlevenshtein,.13)
  (l2levenshtein,.13)
  (cdistlevenshteinn,.13)
  (l2levenshteinn,.13)  
  (cdistapted,.13)
  (l2apted,.13) 
  (cdistaptedn,.13)
  (l2aptedn,.13) 
}; 

\addplot+[draw=black,] plot[error bars/.cd,y dir=both,y explicit relative,] coordinates {
(cdistlevenshtein,0.016953) 
(l2levenshtein,0.059564) 
(cdistlevenshteinn,0.016156) 
(cdistapted,0.016784) 
(l2apted,0.055669) 
(cdistaptedn,0.012874) 
}; 

\addplot+[draw=black,] plot[error bars/.cd,y dir=both,y explicit relative,] coordinates {
(cdistlevenshtein,0.031070) 
(l2levenshtein,0.066374) 
(cdistlevenshteinn,0.012583) 
(l2levenshteinn,-0.046695) 
(cdistapted,0.029735) 
(l2apted,0.058534) 
(cdistaptedn,0.012920) 

}; 

\addplot+[draw=black,] plot[error bars/.cd,y dir=both,y explicit relative,] coordinates {
(cdistlevenshtein,0.047099) 
(l2levenshtein,0.097953) 
(l2levenshteinn,-0.016312) 
(cdistapted,0.043644) 
(l2apted,0.091924) 
}; 

\addplot+[draw=black,] plot[error bars/.cd,y dir=both,y explicit relative,] coordinates {
(cdistlevenshtein,0.021333) 
(l2levenshtein,0.107320) 
(l2levenshteinn,-0.026217) 
(cdistapted,0.018925) 
(l2apted,0.100896) 

}; 

\end{axis}
\end{tikzpicture}}
\subfloat[\label{fig:mfc-def:par}Controlling for paraphrases]{
\begin{tikzpicture}
\begin{axis}[
    ybar,
    width=0.55\textwidth,
    height=4.2cm,
    bar width=.11cm,
    ytick={-0.05 ,-0.025, 0.0, 0.025, 0.05, 0.075, 0.1, 0.125, 0.15, 0.175},
    yticklabels={},
    ymin=-0.075, ymax=0.2,
    ytick align=outside,
    ymajorgrids=true,
    grid style=dashed,
    symbolic x coords={cdistlevenshtein,l2levenshtein,cdistlevenshteinn,l2levenshteinn,cdistapted,l2apted,cdistaptedn,l2aptedn},
    xtick=data, xtick pos=bottom,
    xticklabels={
        $\cos$/Lev., $l_2$/Lev., $\cos$/Lev. n., $l_2$/Lev. n., 
        $\cos$/\textsc{ted}, $l_2$/\textsc{ted}, $\cos$/\textsc{ted} n., $l_2$/\textsc{ted} n.,
        },
    ticklabel style = {font=\footnotesize},
        x tick label style = {rotate=25},
        y tick label style = {font=\tiny},
        typeset ticklabels with strut,
        major y tick style={
            /pgfplots/major tick length=1.5mm,
        },
        nodes near coords always on top/.style={
            scatter/position=absolute,
            positive value/.style={
                at={(axis cs:\pgfkeysvalueof{/data point/x},\pgfkeysvalueof{/data point/y})},
            },
            negative value/.style={
                at={(axis cs:\pgfkeysvalueof{/data point/x},0)},
            },
            every node near coord/.append style={
                check values/.code={%
                    \begingroup
                    \pgfkeys{/pgf/fpu}%
                    \pgfmathparse{\pgfplotspointmeta<0}%
                    \global\let\result=\pgfmathresult
                    \endgroup
                    %
                    %
                    \pgfmathfloatcreate{1}{1.0}{0}%
                    \let\ONE=\pgfmathresult
                    \ifx\result\ONE
                        \pgfkeysalso{/pgfplots/negative value}%
                    \else
                        \pgfkeysalso{/pgfplots/positive value}%
                    \fi
                },
                check values,
                anchor=west,
                rotate=90,
                font=\tiny,
                /pgf/number format/fixed,
                /pgf/number format/zerofill,
                /pgf/number format/precision=3,
                xshift=-0.5ex,
                color=black,
            },
        },
        nodes near coords={
            \pgfmathprintnumber[fixed zerofill,precision=3]{\pgfplotspointmeta}
        },
        nodes near coords always on top,
        cycle list/RdYlBu-5,
        every axis plot/.append style={
            fill,
        },
    ]

\addplot [white,only marks,forget plot, nodes near coords={}, nodes near coords style={}] coordinates {
  (cdistlevenshtein,.13)
  (l2levenshtein,.13)
  (cdistlevenshteinn,.13)
  (l2levenshteinn,.13)  
  (cdistapted,.13)
  (l2apted,.13) 
  (cdistaptedn,.13)
  (l2aptedn,.13) 
}; 

\addplot+[draw=black,error bars/.cd,y dir=both,y explicit relative,] coordinates {
(cdistlevenshtein,0.026041) 
(l2levenshtein,0.051679) 
(cdistlevenshteinn,0.006381) 
(cdistapted,0.026653) 
(l2apted,0.049390) 
(cdistaptedn,0.009925) 
}; 

\addplot+[draw=black,error bars/.cd,y dir=both,y explicit relative,] coordinates {
(cdistlevenshtein,0.044123) 
(l2levenshtein,0.045537) 
(l2levenshteinn,-0.057302) 
(cdistapted,0.043239) 
(l2apted,0.038124) 
(cdistaptedn,0.007234) 
(l2aptedn,-0.028366) 
}; 

\addplot+[draw=black,error bars/.cd,y dir=both,y explicit relative,] coordinates {
(cdistlevenshtein,0.057019) 
(l2levenshtein,0.095568) 
(cdistlevenshteinn,-0.019701) 
(l2levenshteinn,-0.026380) 
(cdistapted,0.054264) 
(l2apted,0.092021) 
}; 

\addplot+[draw=black,error bars/.cd,y dir=both,y explicit relative,] coordinates {
(cdistlevenshtein,0.031095) 
(l2levenshtein,0.097724) 
(cdistlevenshteinn,-0.011062) 
(l2levenshteinn,-0.031973) 
(cdistapted,0.029194) 
(l2apted,0.093480) 
};  

\end{axis}
\end{tikzpicture}}

%% file: figs/tex/figure_meaningdistances_SICK.tex
\begin{tikzpicture}
\begin{axis}[
    ybar,
    width=\textwidth/3,
    height=4cm,
    bar width=.11cm,
    legend style={at={(1.1,0.5)},
      anchor=west,legend columns=-1,
      /tikz/every even column/.append style={column sep=0.5cm}},
    ylabel={\footnotesize Correlation},
    ytick={-1.0,-0.9,..., 0.1},
    yticklabels={-1.00, ,-.80, ,-.60, ,-.40, ,-.20, ,0.00},
    ymin=-1.1, ymax=0.05,
    ytick align=outside,
    ymajorgrids=true,
    grid style=dashed,
    symbolic x coords={cdist,l2,},%
    xtick=data, xtick pos=bottom,
    enlarge x limits=0.5,
    xticklabels={Cosine,Euclidean},
    ticklabel style = {font=\footnotesize},
    y tick label style = {font=\tiny},
    x tick label style = {font=\footnotesize},
    legend style = {font=\small, legend columns=1},
    nodes near coords,
    nodes near coords style={font=\tiny,rotate=90,anchor=east,xshift=.625ex,/pgf/number format/fixed,/pgf/number format/zerofill,/pgf/number format/precision=2, color=black},
    cycle list/RdYlBu-5,
    every axis plot/.append style={
        fill,
    },
    ]







\addplot+[draw=black,] coordinates {
     (cdist, -0.04291815671287734)
     (l2,  -0.22311677052282675)
    };
\addlegendentry{\small random Transf.};

    \addplot+[draw=black,] coordinates {
     (cdist, -0.7689371705824154)
     (l2, -0.7689371357117684)
    };
\addlegendentry{\small \textsc{use}};    
    
\addplot+[draw=black,] coordinates {
     (cdist, -0.4280442076131226)
     (l2,  -0.4187232838753729)
    };
\addlegendentry{\small random \textsc{lstm}};

    \addplot+[draw=black,] coordinates {
      (cdist,  -0.6426971169589213)
      (l2,  -0.6613525906381208)
    };
\addlegendentry{\small Infersent};
    
    \addplot+[draw=black,] coordinates {
     (cdist, -0.48760825123193224)
     (l2,  -0.48760835869009567)
    };
\addlegendentry{\small Skip-Thought};
    

\end{axis}
\end{tikzpicture}

%% file: figs/tex/figure_sents-all.tex
\subfloat[\label{fig:mfc-sent:no-ctrl}No control]{
\begin{tikzpicture}
    \begin{axis}[
        ybar,
        width=0.35\textwidth,
        height=4cm,
        bar width=0.11cm,
        ymin=-0.525, ymax=1.215,
        ytick align=outside,
        ylabel={
            \footnotesize Correlation
        },
        ytick={
            -0.5,-0.375,...,1.125
        },
        yticklabels={
            -.50,,-.25,,.00,,.25,,.50,,.75,,1.00,
        },
        ytick align=outside,
        ymajorgrids=true,
        grid style=dashed,
        xtick=data,
        xtick pos=bottom,
        symbolic x coords={
            cdistlevenshtein,
            l2levenshtein,
            cdistlevenshteinn,
            l2levenshteinn
        },
        xticklabels={
            $\cos$/Lev., $l_2$/Lev., $\cos$/Lev. n., $l_2$/Lev. n., 
        },
        ticklabel style = {font=\footnotesize},
        x tick label style = {rotate=25},
        y tick label style = {font=\tiny},
        typeset ticklabels with strut,
        major y tick style={
            /pgfplots/major tick length=1.5mm,
        },
        nodes near coords always on top/.style={
            scatter/position=absolute,
            positive value/.style={
                at={(axis cs:\pgfkeysvalueof{/data point/x},\pgfkeysvalueof{/data point/y})},
            },
            negative value/.style={
                at={(axis cs:\pgfkeysvalueof{/data point/x},0)},
            },
            every node near coord/.append style={
                check values/.code={%
                    \begingroup
                    \pgfkeys{/pgf/fpu}%
                    \pgfmathparse{\pgfplotspointmeta<0}%
                    \global\let\result=\pgfmathresult
                    \endgroup
                    %
                    %
                    \pgfmathfloatcreate{1}{1.0}{0}%
                    \let\ONE=\pgfmathresult
                    \ifx\result\ONE
                        \pgfkeysalso{/pgfplots/negative value}%
                    \else
                        \pgfkeysalso{/pgfplots/positive value}%
                    \fi
                },
                check values,
                anchor=west,
                rotate=90,
                font=\tiny,
                /pgf/number format/fixed,
                /pgf/number format/zerofill,
                /pgf/number format/precision=3,
                xshift=-0.625ex,
                color=black,
            },
        },
        nodes near coords={
            \pgfmathprintnumber[fixed zerofill,precision=3]{\pgfplotspointmeta}
        },
        nodes near coords always on top,
        legend style={at={(0.5,1.1)},
            anchor=south,legend columns=-1,
            /tikz/every even column/.append style={column sep=0.5cm}
        },
        cycle list/RdYlBu-5,
        every axis plot/.append style={
            fill,
        },
    ]


\addplot+[draw=black,error bars/.cd,y dir=both,y explicit relative,] coordinates {

(cdistlevenshtein,-0.503818) 
(l2levenshtein,0.753789) 
(cdistlevenshteinn,-0.056227) 
(l2levenshteinn,0.238361) 

};

\addplot+[draw=black,error bars/.cd,y dir=both,y explicit relative,] coordinates {
(cdistlevenshtein,0.113231) 
(l2levenshtein,0.113984) 
(cdistlevenshteinn,0.342222) 
(l2levenshteinn,0.345006) 
};

\legend{{\footnotesize{random Transf.}},{\footnotesize{\textsc{use}}}};

\end{axis}
\end{tikzpicture}}
\subfloat[\label{fig:mfc-sent:isp}Controlling for stop-words]{
\begin{tikzpicture}
\begin{axis}[
        ybar,
        width=0.33\textwidth,
        height=4cm,
        bar width=0.11cm,
        ymin=-0.525, ymax=1.215,
        ytick align=outside,
        ylabel={
        },
        ytick={
            -0.5,-0.375,...,1.125
        },
        yticklabels={
        },
        ytick align=outside,
        ymajorgrids=true,
        grid style=dashed,
        xtick=data,
        xtick pos=bottom,
        symbolic x coords={
            cdistlevenshtein,
            l2levenshtein,
            cdistlevenshteinn,
            l2levenshteinn
        },
        xticklabels={
            $\cos$/Lev., $l_2$/Lev., $\cos$/Lev. n., $l_2$/Lev. n., 
        },
        ticklabel style = {font=\footnotesize},
        x tick label style = {rotate=25},
        y tick label style = {font=\footnotesize},
        typeset ticklabels with strut,
        major y tick style={
            /pgfplots/major tick length=1.5mm,
        },
        nodes near coords always on top/.style={
            scatter/position=absolute,
            positive value/.style={
                at={(axis cs:\pgfkeysvalueof{/data point/x},\pgfkeysvalueof{/data point/y})},
            },
            negative value/.style={
                at={(axis cs:\pgfkeysvalueof{/data point/x},0)},
            },
            every node near coord/.append style={
                check values/.code={%
                    \begingroup
                    \pgfkeys{/pgf/fpu}%
                    \pgfmathparse{\pgfplotspointmeta<0}%
                    \global\let\result=\pgfmathresult
                    \endgroup
                    %
                    %
                    \pgfmathfloatcreate{1}{1.0}{0}%
                    \let\ONE=\pgfmathresult
                    \ifx\result\ONE
                        \pgfkeysalso{/pgfplots/negative value}%
                    \else
                        \pgfkeysalso{/pgfplots/positive value}%
                    \fi
                },
                check values,
                anchor=west,
                rotate=90,
                font=\tiny,
                /pgf/number format/fixed,
                /pgf/number format/zerofill,
                /pgf/number format/precision=3,
                xshift=-0.625ex,
                color=black,
            },
        },
        nodes near coords={
            \pgfmathprintnumber[fixed zerofill,precision=3]{\pgfplotspointmeta}
        },
        nodes near coords always on top,
        cycle list/RdYlBu-5,
        every axis plot/.append style={
            fill,
        },
    ]


\addplot+[draw=black,error bars/.cd,y dir=both,y explicit relative,] coordinates {

(cdistlevenshtein,-0.480549) 
(l2levenshtein,0.648706) 
(cdistlevenshteinn,-0.033080) 
(l2levenshteinn,0.087830) 
};

\addplot+[draw=black,error bars/.cd,y dir=both,y explicit relative,] coordinates {
(cdistlevenshtein,0.089955) 
(l2levenshtein,0.091419) 
(cdistlevenshteinn,0.306058) 
(l2levenshteinn,0.311477) 
};

\end{axis}
\end{tikzpicture}}
\subfloat[\label{fig:mfc-sent:syn}Controlling for synonymy]{
\begin{tikzpicture}
    \begin{axis}[
        ybar,
        width=0.33\textwidth,
        height=4cm,
        bar width=0.11cm,
        ymin=-0.525, ymax=1.215,
        ytick align=outside,
        ylabel={
        },
        ytick={
            -0.5,-0.375,...,1.125
        },
        yticklabels={
        },
        ytick align=outside,
        ymajorgrids=true,
        grid style=dashed,
        xtick=data,
        xtick pos=bottom,
        symbolic x coords={
            cdistlevenshtein,
            l2levenshtein,
            cdistlevenshteinn,
            l2levenshteinn
        },
        xticklabels={
            $\cos$/Lev., $l_2$/Lev., $\cos$/Lev. n., $l_2$/Lev. n., 
        },
        ticklabel style = {font=\footnotesize},
        x tick label style = {rotate=25},
        y tick label style = {font=\footnotesize},
        typeset ticklabels with strut,
        major y tick style={
            /pgfplots/major tick length=1.5mm,
        },
        nodes near coords always on top/.style={
            scatter/position=absolute,
            positive value/.style={
                at={(axis cs:\pgfkeysvalueof{/data point/x},\pgfkeysvalueof{/data point/y})},
            },
            negative value/.style={
                at={(axis cs:\pgfkeysvalueof{/data point/x},0)},
            },
            every node near coord/.append style={
                check values/.code={%
                    \begingroup
                    \pgfkeys{/pgf/fpu}%
                    \pgfmathparse{\pgfplotspointmeta<0}%
                    \global\let\result=\pgfmathresult
                    \endgroup
                    %
                    %
                    \pgfmathfloatcreate{1}{1.0}{0}%
                    \let\ONE=\pgfmathresult
                    \ifx\result\ONE
                        \pgfkeysalso{/pgfplots/negative value}%
                    \else
                        \pgfkeysalso{/pgfplots/positive value}%
                    \fi
                },
                check values,
                anchor=west,
                rotate=90,
                font=\tiny,
                /pgf/number format/fixed,
                /pgf/number format/zerofill,
                /pgf/number format/precision=3,
                xshift=-0.625ex,
                color=black,
            },
        },
        nodes near coords={
            \pgfmathprintnumber[fixed zerofill,precision=3]{\pgfplotspointmeta}
        },
        nodes near coords always on top,
        cycle list/RdYlBu-5,
        every axis plot/.append style={
            fill,
        },
    ]
    
\addplot+[draw=black,error bars/.cd,y dir=both,y explicit relative,] coordinates {

(cdistlevenshtein,-0.503872) 
(l2levenshtein,0.753458) 
(cdistlevenshteinn,-0.055141) 
(l2levenshteinn,0.232666) 

};

\addplot+[draw=black,error bars/.cd,y dir=both,y explicit relative,] coordinates {

(cdistlevenshtein,0.113834) 
(l2levenshtein,0.114588) 
(cdistlevenshteinn,0.345460) 
(l2levenshteinn,0.348280) 

};


    \end{axis}
\end{tikzpicture}
}